\documentclass[letterpaper]{article} 
\usepackage{aaai2026}  
\usepackage{times}  
\usepackage{helvet}  
\usepackage{courier}  
\usepackage[hyphens]{url}  
\usepackage{graphicx} 
\usepackage{natbib}  
\usepackage{caption} 
\usepackage{algorithm}
\usepackage{algorithmic}

\usepackage{url}
\usepackage{booktabs}
\usepackage[most]{tcolorbox}
\usepackage{amsmath}
\usepackage{multirow}
\usepackage{graphicx}
\usepackage{amssymb}
\usepackage{xcolor}

\usepackage{cleveref}
\usepackage{newfloat}
\usepackage{listings}
\DeclareCaptionStyle{ruled}{labelfont=normalfont,labelsep=colon,strut=off} 
\floatstyle{ruled}
\newfloat{listing}{tb}{lst}{}
\floatname{listing}{Listing}
\title{MathSmith: Towards Extremely Hard Mathematical Reasoning by Forging Synthetic Problems with a Reinforced Policy}
\author{
    Shaoxiong Zhan\textsuperscript{\rm 1}, 
    Yanlin Lai\textsuperscript{\rm 1}, 
    Ziyu Lu\textsuperscript{\rm 1}, 
    Dahua Lin\textsuperscript{\rm 2}, 
    Ziqing Yang\textsuperscript{\rm 3}\thanks{Corresponding authors.}, 
    Fei Tan\textsuperscript{\rm 4}\footnotemark[1]
}
\affiliations{
    \textsuperscript{\rm 1}Tsinghua University\\
    \textsuperscript{\rm 2}The Chinese University of Hong Kong\\
    \textsuperscript{3}SenseTime Research\\
    \textsuperscript{4}East China Normal University\\
    \{zhansx24, laiyl24, luziyu24\}@mails.tsinghua.edu.cn, \\dhlin@ie.cuhk.edu.hk, ziqingyang@gmail.com, ftan@mail.ecnu.edu.cn
}

\begin{document}

\maketitle

\begin{abstract}
Large language models have achieved substantial progress in mathematical reasoning, yet their advancement is limited by the scarcity of high-quality, high-difficulty training data. Existing synthesis methods largely rely on transforming human-written templates, limiting both diversity and scalability. We propose MathSmith, a novel framework for synthesizing challenging mathematical problems to enhance LLM reasoning. Rather than modifying existing problems, MathSmith constructs new ones from scratch by randomly sampling concept–explanation pairs from PlanetMath, ensuring data independence and avoiding contamination. To increase difficulty, we design nine predefined strategies as soft constraints during rationales. We further adopts reinforcement learning to jointly optimize structural validity, reasoning complexity, and answer consistency. The length of the reasoning trace generated under autoregressive prompting is used to reflect cognitive complexity, encouraging the creation of more demanding problems aligned with long-chain-of-thought reasoning. Experiments across five benchmarks, categorized as easy \& medium (GSM8K, MATH-500) and hard (AIME2024, AIME2025, OlympiadBench), show that MathSmith consistently outperforms existing baselines under both short and long CoT settings. Additionally, a weakness-focused variant generation module enables targeted improvement on specific concepts. Overall, MathSmith exhibits strong scalability, generalization, and transferability, highlighting the promise of high-difficulty synthetic data in advancing LLM reasoning capabilities. Our code and data are available at \url{https://github.com/Jasaxion/MathSmith}.
\end{abstract}


\begin{figure*}[htbp]
\centering
\includegraphics[width=0.94\textwidth]{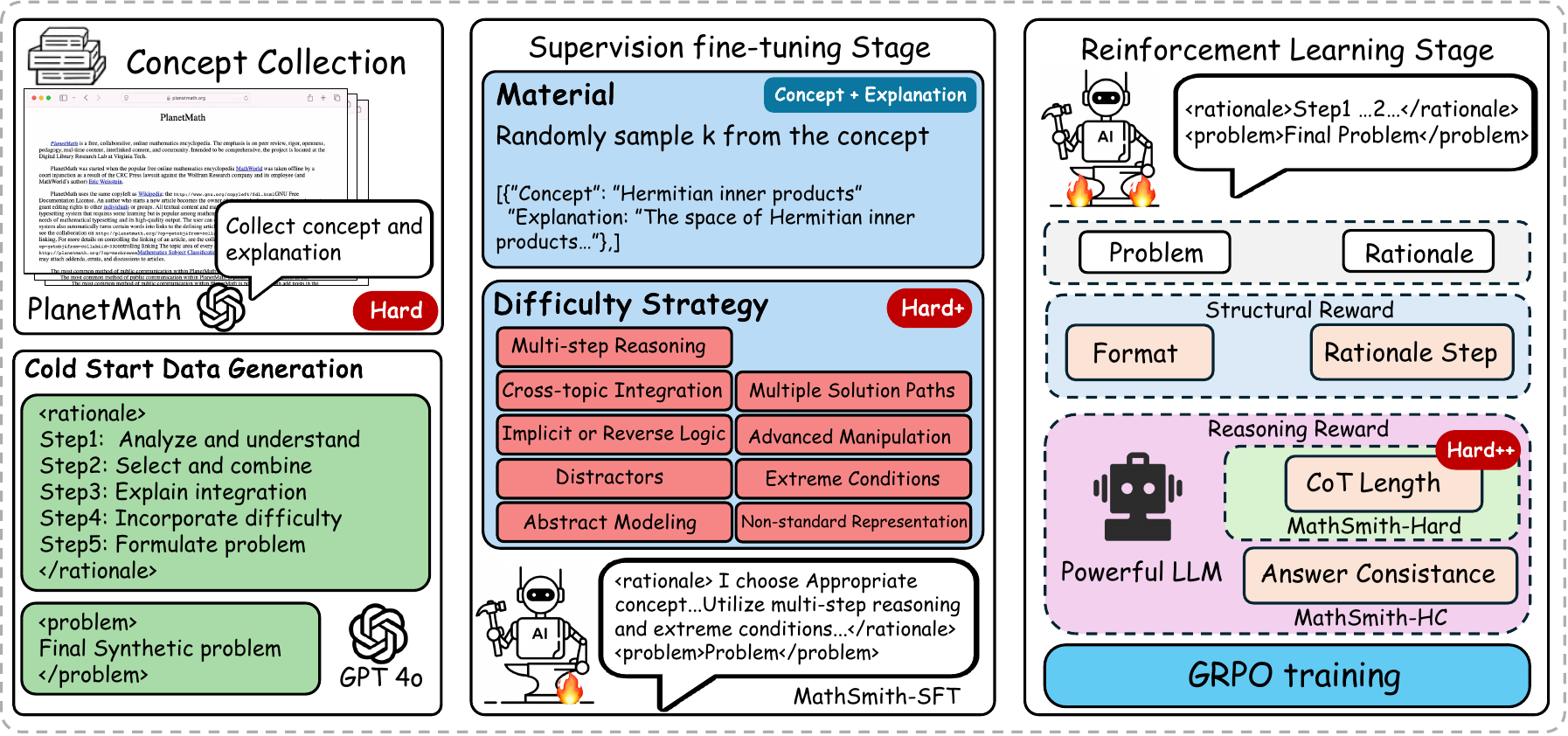}
\caption{The MathSmith workflow, comprising the phases of concept and explanation collection, supervised fine-tuning, and reinforcement learning.}
\label{mathsmith_framework}
\end{figure*}

\section{Introduction}
\label{intro}

In recent years, large language models (LLMs) have achieved remarkable progress in reasoning tasks across mathematics, science, and programming~\cite{guo2025deepseek, jaech2024openai, zhang2024balancing, gu2024cmr, ding2025consultant}. As models continue to scale in both size and architectural sophistication, their capabilities have expanded from solving elementary-level math problems to addressing Olympiad-level challenges~\cite{shao2024deepseekmath, openai2025systemcard, lu2023makes}, gradually revealing their potential for general-purpose intelligence. However, the advancement of reasoning ability now faces a critical bottleneck: the scarcity of high-quality, high-difficulty mathematical problems for training and evaluation limits the upper bound of model performance. Moreover, a lack of diversity in existing problem distributions raises concerns about models memorizing recurring patterns rather than truly reasoning~\cite{huang2025mathperturb, zhao2025more}, further highlighting the urgent need for diverse and challenging mathematical data to support continued progress.

Most existing approaches to mathematical problem synthesis rely on extracting templates, structures, or conceptual patterns from existing questions~\cite{huang2025key}, followed by rewriting~\cite{yu2024metamath}, augmentation~\cite{toshniwal2025openmathinstruct, liu2025augmenting}, question back-translation~\cite{lu2024mathgenie}, or evolutionary transformations~\cite{luo2023wizardmath}. While these methods enhance data diversity to some extent, they remain fundamentally constrained by the distribution and structure of human-authored problems, often lacking generation autonomy and precise difficulty control. As articulated in the Bitter Lesson~\cite{sutton2019bitter}, sustainable progress in AI is ultimately driven by general purpose, computation-heavy methods rather than handcrafted knowledge. In line with this perspective, we argue that future reasoning agents should be able to autonomously generate high-quality, intellectually challenging problems. To this end, we introduce \textbf{MathSmith}, a novel framework that emulates the role of a mathematical blacksmith: it extracts raw materials (i.e., concept and explanation pairs) and progressively refines them into complex and coherent mathematical problems.

\begin{figure*}[htbp]
\centering
\includegraphics[width=0.8\textwidth]{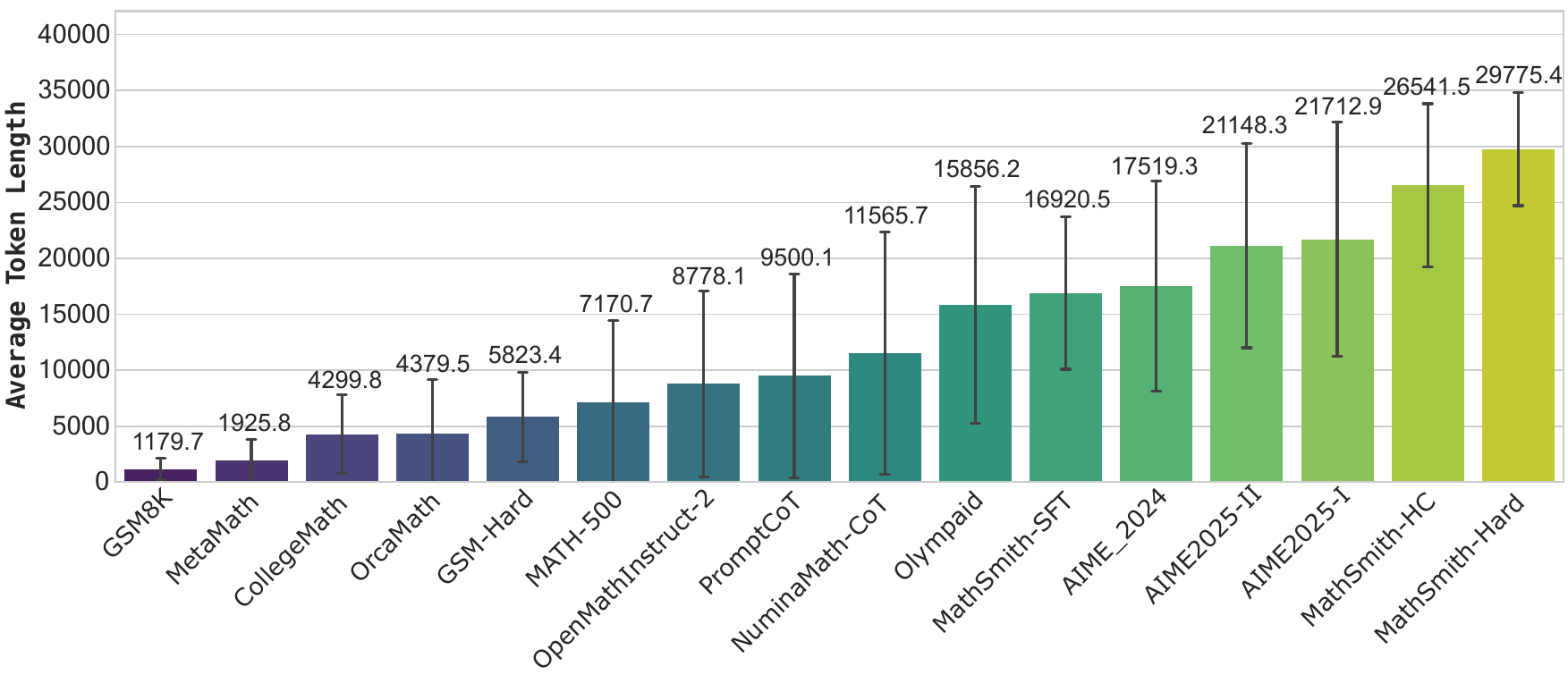}
\caption{Average reasoning trace token length under thinking mode (Qwen3-30B-A3B) across various open-source math datasets. Problems synthesized by MathSmith variants elicit significantly longer reasoning, reflecting higher complexity.}
\label{cotlength-cal}
\end{figure*}

To enhance both the difficulty and the quality of the synthesized problems, we begin by analyzing the structural and cognitive features of existing high-difficulty questions. This analysis leads to the formulation of nine pre-defined difficulty strategies in Figure~\ref{mathsmith_framework}, including multi-step reasoning, cross-topic integration, implicit or reverse logic, distractor construction, abstract modeling, multiple solution paths, advanced manipulation extreme conditions and non-standard representation, which serve as soft constraints during generation. Additionally, we introduce a reinforcement learning stage to optimize the generation process across three dimensions: structural validity, reasoning complexity, and answer consistency. Drawing inspiration from the long CoT paradigm~\cite{guo2025deepseek, wang2024reward}, we adopt the reasoning trace length, as produced by models under autoregressive CoT prompting, as an indirect estimation of problem difficulty, and incorporate it into our reward design. We observe that more challenging problems tend to elicit significantly longer reasoning traces, as shown in Figure~\ref{cotlength-cal}, suggesting a potential link between problem difficulty and reasoning depth. While it remains uncertain whether such problems definitively enhance the reasoning capabilities of LLMs, they offer a promising direction worth exploring. Building on this insight, our method synthesizes problems that induce longer reasoning sequences, enriching the pool of high-difficulty data and fostering deeper reasoning in LLMs.

We categorize widely used benchmarks into two difficulty tiers: easy \& medium (GSM8K, MATH-500) and hard (AIME2024, AIME2025, Olympiad). Compared to a variety of existing methods, MathSmith achieves significantly better performance under both short-CoT and long-CoT prompt settings, producing relative improvements of 9.8\%–18.1\% on the hard benchmarks. In addition, our weakness-focused variant generation mechanism effectively improves model performance on specific underperforming concepts, and the synthesized problems generalize well across different reasoning tasks. Moreover, extended experiments show that MathSmith maintains strong performance as both the number of problems and the model size increase, demonstrating its superiority in reasoning depth, scalability, and effectiveness on larger models.

Our main contributions are as follows: (1) We synthesize mathematical problems by randomly sampling concept–explanation pairs and constructing problems through step-by-step rationale generation, avoiding reliance on real-world templates and minimizing data contamination; (2) We propose the MathSmith framework, which incorporates nine difficulty strategies, a multi-objective reinforcement learning mechanism for optimizing structural integrity, reasoning depth, and solution consistency, as well as a weakness-focused variant generation module for targeted concept-level improvement; (3) We conduct extensive evaluations demonstrating that the synthesized problems significantly enhance model performance on challenging benchmarks such as AIME2024, AIME2025, and Olympiad, particularly under long-chain-of-thought prompting setups.

\section{Related Work}
\label{related_work}

\textbf{Mathematical Reasoning with Large Language Models.}  
Large language models (LLMs) have shown growing capabilities in mathematical reasoning, driven by both training and inference advancements. At the training stage, some works enhance mathematical foundations through continued pre-training on large-scale corpora~\cite{wang2024mathpile, du2025resure, zhang-etal-2025-dadpo}, while others focus on post-training with curated instruction datasets such as OpenMathInstruct~\cite{toshniwal2025openmathinstruct}, Mammoth~\cite{yue2024mammoth} and DART-Math~\cite{tong2024dart}. For inference, Chain-of-Thought (CoT) prompting~\cite{wei2022chain} enables step-by-step reasoning, and Program-of-Thought (PoT)~\cite{chen2023program} incorporates the use of tools for solving complex problems. PromptCOT~\cite{zhao2025promptcot} combines concept-driven prompts with multi-step planning to generate Olympiad-level problems and is most relevant to our work, though it still relies on human-selected concepts and lacks deeper reasoning control. Other approaches, such as Llemma~\cite{azerbayev2024llemma} and MathPrompter~\cite{imani2023mathprompter}, further explore structured or task-specific alignment. In contrast, our method controls the reasoning structure from the source, targeting logical consistency and difficulty through both supervised and reinforcement training.

\textbf{Math Instruction Synthesis and Difficulty Control.}  
Recent studies have explored data synthesis methods to improve LLMs on mathematical tasks. MetaMath~\cite{yu2024metamath} increases the diversity of problems through rewriting, NuminaMath~\cite{li2024numinamath} resamples the benchmarks with CoT guidance, and GSM-Plus~\cite{li-etal-2024-gsm} increases the robustness with distribution-based augmentation, while others like ScaleQuest~\cite{ding2024unleashing} focus on generating novel questions from scratch. To better control difficulty, JiuZhang3.0~\cite{zhou2024jiuzhang3} structures prompts tiered by educational stage. PromptCOT, MathScale~\cite{tang2024mathscale} and WizardMath~\cite{luo2023wizardmath} introduce concept planning, graph-based concept combination, and reinforcement-driven evolution to guide generation. Key-point-driven methods~\cite{huang2025key} further support multi-concept integration. However, most methods still rely on seed prompts derived from real human-written math problems, limiting their generative autonomy. Moreover, their difficulty control typically depends on prompt-level labels assigned by language models, which lack objective justification. Our framework constructs problems from randomly sampled concept–explanation pairs, bypassing existing problems entirely. With structural, complexity, and consistency-aware rewards, MathSmith generates high-quality, verifiable problems that push LLMs toward stronger mathematical reasoning.

\section{Methodology}
\label{method}

As illustrated in Figure~\ref{mathsmith_framework}, the MathSmith framework generates challenging mathematical problems through a core three-stage process: (1)~\textbf{Concept-Explanation Collection}: collecting challenging ``Concept + Explanation'' pairs from PlanetMath; (2)~\textbf{Supervised Fine-Tuning Stage}: employing SFT on a seed dataset generated by GPT-4o to equip the model with an initial reasoning ability for formatted problem generation; and (3)~\textbf{Reinforcement Learning Stage}: implementing RL to refine problem difficulty, where a reward function combines signals from format, solution complexity, and answer consistency.

\begin{figure}[htbp]
\centering
\includegraphics[width=0.93\columnwidth]{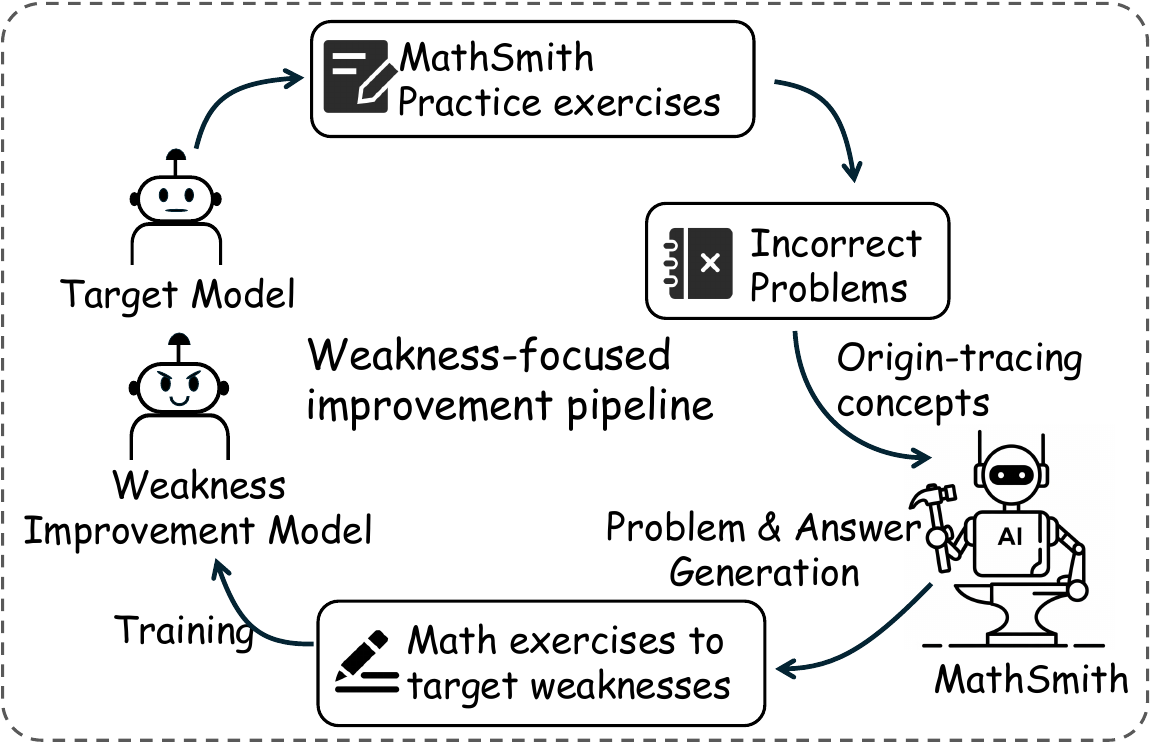}
\caption{The MathSmith weakness-focused improvement pipeline. Problems are traced to concept explanations, which serve as a basis for generating targeted variants to strengthen model weaknesses.}
\label{self-weakness_pipeline}
\end{figure}

Furthermore, the traceability of MathSmith-synthesized problems to their source concepts enables \textbf{Weakness-Focused Improvement Pipeline}, a module for targeted enhancement of model weaknesses, as shown in Figure~\ref{self-weakness_pipeline}.

\subsection{Concept and Explanation Collection}
\label{concept-exp-collection}

We construct a dataset that contains 11,000 mathematical concepts and their explanations. The data for this dataset is sourced from PlanetMath~\cite{planetmath}, a repository known for its extensive coverage of advanced mathematics and theoretically deep concepts. This choice ensures our resulting collection of concepts is inherently challenging. We first crawl the mathematics-related pages from its website and filter out entries unrelated to mathematical concepts to ensure a clear conceptual focus. Subsequently, we utilize the GPT-4o to automatically summarize the core concept of each page, which generates a collection of ``Concept + Explanation'' pairs.

\subsection{Supervise Fine-tuning Stage}
\label{sft-stage}

We adopt Qwen3-8B as the base model for problem generation and use GPT-4o to synthesize cold-start training data. Specifically, we randomly sample five concepts and their corresponding explanations from the constructed concept collection and provide them to the model as seed inputs, prompting it to generate math problems grounded in the given instructions and aligned with the sampled concepts.

Each generated sample is structured into two components: a rationale section that outlines the problem construction process, and a problem section that presents the final question. As illustrated in the ``Cold Start Data Generation'' module of Figure~\ref{mathsmith_framework}, the rationale consists of exactly five reasoning steps. This format serves both as a pedagogical scaffold and as a structural constraint during training, ensuring consistency in problem synthesis.


To further enhance the difficulty of synthesized problems and encourage advanced mathematical reasoning, we incorporate insights from existing challenging Olympiad problems. Based on these, we design nine predefined \textit{difficulty strategies}, also detailed in Figure~\ref{mathsmith_framework}. Each generated problem is required to incorporate at least two strategies to ensure sufficient complexity. Following this process, we generate about 8k cold-start samples to fine-tune Qwen3-8B, resulting in \textbf{MathSmith-SFT}.

\subsection{Reinforcement Learning Stage}
\label{rl-stage}

We design a composite reward to guide the policy model toward generating valid, challenging, and consistent mathematical problems, comprising structural, complexity, and consistency components.

\paragraph{(1) Structural Reward.}
We assess whether the output contains both ``rationale'' and ``problem'' segments using a binary reward $r_{\text{format}} \in \{0,1\}$. We further compute a step count reward $r_{\text{step}}$ based on the number of reasoning steps $N_{\text{step}}$ in the rationale:
\begin{equation}
r_{\text{step}} =
\begin{cases}
\frac{N_{\text{step}}}{5}, & N_{\text{step}} \le 5, \\
\max\left(1 - \frac{N_{\text{step}} - 5}{5},\, 0\right), & \text{otherwise}.
\end{cases}
\end{equation}

\begin{equation}
r_{\text{structure}}
= \alpha_{\text{format}}\cdot r_{\text{format}}
+ \alpha_{\text{step}}\cdot r_{\text{step}}.
\label{structural_reward}
\end{equation}
Here, $\alpha_{\text{format}}$ and $\alpha_{\text{step}}$ are weighting coefficients. The reward is maximized when the rationale contains exactly five steps, aligning with the prompt template.

\paragraph{(2) Reasoning Complexity Reward.}
To evaluate the difficulty of each generated problem, we utilize the teacher model Qwen3-30B-A3B to generate solutions. The complexity is estimated by measuring the token length of its reasoning trace.

Let $\ell^{(i)}_{\text{cot}}$ be the token length of the $i$-th reasoning trace among $K$ independent samples. The complexity reward is computed as:
\begin{equation}
r_{\text{complexity}} = \frac{1}{K \cdot T_{\text{max}}} \sum_{i=1}^{K} \ell^{(i)}_{\text{cot}},\quad r_{\text{complexity}} \in [0, 1]
\end{equation}
where $T_{\text{max}}$ is a normalization constant (i.e., the maximum allowed CoT length).

\paragraph{Motivation for Reasoning-based Reward.}
We adopt reasoning trace length as a heuristic measure of problem complexity. Intuitively, more challenging problems tend to require deeper and more structured reasoning, leading to longer CoT traces under autoregressive prompting. While longer traces do not directly imply better generalization, they often contain low-entropy intermediate tokens~\cite{wang2025beyond}, which are shown to provide more informative supervision signals during training. Thus, we encourage synthesis of problems that induce longer reasoning as a proxy for higher cognitive complexity and better transferability.

\paragraph{(3) Answer Consistency Reward.}
To evaluate solution consistency, we sample $K$ answers $\mathcal{A} = \{a_1, \dots, a_K\}$ from the teacher model. If a majority answer exists (i.e., $\exists\, a \in \mathcal{A}\ \text{s.t.}\ \mathrm{count}(a) > K/2$), we assign a reward of 1; otherwise, 0:
\begin{equation}
r_{\text{consistency}} =
\begin{cases}
1, & \text{if majority answer exists}, \\
0, & \text{otherwise}.
\end{cases}
\end{equation}
This encourages the policy model to generate problems that elicit deterministic reasoning behavior from the teacher model, indicating clarity and unambiguity in problem formulation.

\paragraph{(4) Final Reward.}
We combine the complexity and consistency objectives into a unified reasoning-oriented reward:
\begin{equation}
\label{reasoning_reward}
r_{\text{reasoning}} = \beta_{\text{complexity}} \cdot r_{\text{complexity}} + \beta_{\text{consistency}} \cdot r_{\text{consistency}},
\end{equation}
where $\beta_{\text{complexity}}$ and $\beta_{\text{consistency}}$ are weighting coefficients.

The reinforcement signal provided to the policy model is the sum of structural and reasoning-based components:
\begin{equation}
\label{eq:total reward}
r_{\text{total}} = r_{\text{structure}} + r_{\text{reasoning}}.
\end{equation}
This reward guides the policy model to generate mathematically valid, non-trivial and verifiable problems by combining structural alignment, reasoning complexity, and answer consistency. We refer to the resulting model trained with both complexity and consistency components as \textbf{MathSmith-HC}, while the variant that excludes the consistency term and uses only the complexity reward is referred to as \textbf{MathSmith-Hard}.

We employ Group Relative Policy Optimization (GRPO) to optimize the policy model $\pi_\theta$, maximizing the expected final reward. For each input $c$, consisting of a set of five sampled concepts and their corresponding explanation, the policy model is prompted to generate a group of $G$ math problems $\{o_i\}_{i=1}^G$. We then evaluate each problem using our composite reward function Eq.~\eqref{eq:total reward} to obtain a scalar reward $R_i$. The advantage of the $i$-th problem is calculated by normalizing the token-level rewards:

\begin{equation}
\hat{A}_{i,t} = \frac{R_i - \text{mean}(\{R_j\}_{j=1}^G)}{\text{std}(\{R_j\}_{j=1}^G)}.
\end{equation}
GRPO then maximizes the clipped objective:

\begin{equation}
\begin{split}
    \mathcal{J}_{\text{GRPO}}(\theta) &= \mathbb{E}_{c, \{o_i\} \sim \pi_{\theta_{\text{old}}}} \\
    & \left[ \frac{1}{G} \sum_{i=1}^{G} \frac{1}{|o_i|}\sum_{t=1}^{|o_i|}  \mathcal{L}_{i,t} - \beta \mathbb{D}_{KL}(\pi_\theta  \|\pi_{\text{ref}}) \right],
\end{split}
\end{equation}
where
\begin{equation}
    \mathcal{L}_{i,t} = \min \Big( r_{i,t}(\theta) \hat{A}_{i,t}, \text{clip}(r_{i,t}(\theta), 1-\epsilon, 1+\epsilon) \hat{A}_{i,t} \Big), 
\end{equation}
\begin{equation}
    r_{i,t}(\theta) = \frac{\pi_\theta(o_{i,t} | c, o_{i,<t})}{\pi_{\theta_{\text{old}}}(o_{i,t} | c, o_{i,<t})}.
\end{equation}
In these equations, $\pi_\theta$ is the policy before the update, $\pi_{\text{ref}}$ is the reference policy (MathSmith-SFT). The hyperparameters $\epsilon$ and $\beta$ control the clipping threshold and the KL penalty term, respectively. By maximizing this objective, we aim to iteratively improve the model's capability to generate well-structured and inherently challenging problems that demand complex multi-step reasoning.

\subsection{Weakness-focused Improvement Pipeline}
\label{self-weakness-pipeline}

Given that each MathSmith-generated problem is explicitly linked to a concept set, we design a pipeline to enhance model performance on identified weaknesses.

\textbf{Practice Set $Q$:}  
We generate $|Q| = 1000$ problems using the MathSmith generator, covering a broad range of concepts. For each $q \in Q$, we sample 32 completions from Qwen3-30B-A3B, and select the most frequent answer as the reference solution. After filtering, we retain 923 high-quality items.

\textbf{Variant Set $Q'$:}  
For each $q \in Q$ with concept set $c$, we generate a variant problem $q' \sim \mathcal{G}(c)$, where $\mathcal{G}(c)$ denotes the MathSmith generator conditioned on concept $c$. The resulting set $Q'$ forms a variant set aligned with concepts.

We fine-tune the model on a small subset of $Q'$ using supervised learning. Let $\mathcal{M}_\text{base}$ denote the initial model and $\mathcal{M}_\text{imp}$ the fine-tuned model. We iteratively update $\mathcal{M}_\text{base} \rightarrow \mathcal{M}_\text{imp}$ until:
\begin{equation}
\text{Acc}_{Q}(\mathcal{M}_\text{imp}) \geq \tau,
\end{equation}
where $\tau$ is a predefined accuracy threshold on $Q$. The resulting model $\mathcal{M}_\text{imp}$ is referred to as the improved model.

\begin{table*}[htbp]
\centering
\resizebox{0.88\textwidth}{!}{%
\begin{tabular}{clcccccc}
\hline
\multicolumn{1}{c}{\multirow{2}{*}{Models}} & \multicolumn{1}{c}{\multirow{2}{*}{Method}} & \multicolumn{2}{c}{Eeay   \& Medium} & \multicolumn{3}{c}{Hard} & \multirow{2}{*}{Avg for Hard (Rel.   Imp.)} \\
\multicolumn{1}{l}{} & \multicolumn{1}{c}{} & \multicolumn{1}{l}{GSM8K} & \multicolumn{1}{l}{MATH-500} & \multicolumn{1}{l}{AIME2024} & \multicolumn{1}{l}{AIME2025} & \multicolumn{1}{l}{Olympiad} &  \\ \hline
\multicolumn{1}{l}{} & - & \textbf{92.2} & 72.2 & 16.7 & 6.7 & 38.6 & 20.7 \\
\multicolumn{1}{l}{} & MetaMath & 82.6 & 61.0 & 13.3 & 0.0 & 26.3 & 13.2 (-36.1\%) \\
Qwen-2.5-7B-Instruct & NuminaMath-COT & 87.2 & 73.4 & \textbf{23.3} & 3.3 & 36.9 & 21.2  (+2.4\%) \\
short-CoT & OpenMathInstruct-2 & 88.2 & 74.6 & 16.7 & 6.7 & 38.4 & 20.6 (-0.3\%) \\
 & PromptCOT & 87.6 & 73.2 & \textbf{23.3} & 6.7 & 35.9 & 21.9 (+6.2\%) \\
 & MathSmith-HC & 91.2 & \textbf{75.2} & \textbf{23.3} & \textbf{10.0} & \textbf{39.9} & \textbf{24.4 (+18.1\%)} \\ \hline
\multicolumn{1}{l}{} & - & \textbf{93.4} & 82.8 & 30.0 & 16.7 & 51.0 & 32.6 \\
\multicolumn{1}{l}{} & MetaMath & 92.3 & 81.4 & 30.0 & 20.0 & 51.0 & 33.7 (+3.4\%) \\
Qwen-3-8B & NuminaMath-COT & 92.6 & 83.2 & \textbf{33.3} & 13.3 & 50.6 & 32.4 (-0.5\%) \\
short-CoT & OpenMathInstruct-2 & 92.8 & 84.0 & 23.3 & 13.3 & 51.6 & 29.4 (-9.7\%) \\
 & PromptCOT & 92.5 & 83.8 & 26.7 & \textbf{23.3} & 49.9 & 33.3 (+2.3\%) \\
 & MathSmith-HC & 92.9 & \textbf{84.4} & \textbf{33.3} & \textbf{23.3} & \textbf{53.1} & \textbf{36.6 (+12.3\%)} \\ \hline \hline
 & - & 89.3 & 88.6 & 43.3 & 36.7 & 52.4 & 44.1 \\
DS-R1-qwen2.5-7B & Numinamath-COT & 93.3 & 91.0 & 46.7 & 30.0 & 53.4 & 43.4 (-1.7\%) \\
long-CoT & OpenMathInstruct-2 & \textbf{93.9} & 91.2 & 46.7 & 40.0 & 56.1 & 47.6 (+7.9\%) \\
 & PromptCOT & 93.5 & 91.0 & 46.7 & 33.3 & 55.8 & 45.3 (+2.6\%) \\
 & MathSmith-HC & 89.2 & \textbf{91.6} & \textbf{53.3} & \textbf{43.3} & \textbf{56.5} & \textbf{51.0 (+15.6\%)} \\ \hline
 & - & 94.8 & 94.4 & 66.7 & 63.3 & 66.2 & 65.4 \\
Qwen3-8B & Numinamath-COT & \textbf{95.5} & 96.0 & 73.3 & 63.3 & 68.1 & 68.2 (+4.3\%) \\
long-CoT & OpenMathInstruct-2 & \textbf{95.5} & 95.8 & 70.0 & 60.0 & 67.4 & 65.8 (+0.6\%) \\
 & PromptCOT & 95.1 & 95.4 & 73.3 & 63.3 & 67.1 & 67.9 (+3.8\%) \\
 & MathSmith-HC & 95.1 & \textbf{96.4} & \textbf{76.7} & \textbf{70.0} & \textbf{68.8} & \textbf{71.8 (+9.8\%)} \\ \hline
\end{tabular}%
}
\caption{Baseline performance under equal data and training conditions. MathSmith achieves consistently better generalization on challenging problems.}
\label{tab:main_resuilt}
\end{table*}

\section{Experimental Setups}
\label{exps}

\subsection{Datasets and Evaluation Metrics}
\label{dataset_and_eval}

To evaluate mathematical reasoning, we adopt five representative benchmarks categorized into two difficulty tiers: easy \& medium and hard, with all results reported by \textbf{Pass@1}. The former includes (1) \textbf{GSM8K}~\cite{cobbe2021training}, focusing on math word problems, and (2) \textbf{MATH}~\cite{lightman2023let}, covering high school-level reasoning. The hard tier comprises three competition-style benchmarks: (3) \textbf{AIME 2024}\footnote{\url{https://huggingface.co/datasets/Maxwell-Jia/AIME_2024}}, (4) \textbf{AIME 2025}\footnote{\url{https://huggingface.co/datasets/opencompass/AIME2025}}, and (5) \textbf{OlympiadBench}~\cite{he2024olympiadbench}, which require symbolic and multi-step reasoning to solve advanced mathematical problems.

\subsection{Baseline}
\label{baseline}

We compare with four representative math problem generation approaches. (1) \textbf{OpenMathInstruct}~\cite{toshniwal2025openmathinstruct} synthesizes new problems by prompting LLMs with in-context examples, relying on solution extrapolation without explicit control over difficulty. (2) \textbf{NuminaMath}~\cite{li2024numinamath} reformulates seed problems through CoT-guided sampling, aligning generated problems with existing math benchmarks. (3) \textbf{MetaMath}~\cite{yu2024metamath} increases problem diversity through structured rewriting techniques, including inversion, rephrasing, and reverse construction. (4) \textbf{PromptCOT} targets the difficulty level of the Olympiad by conditioning on the mathematical concepts sampled and guiding the generation through multistep rational planning. For each baseline, we sample 50K problems from its official dataset and regenerate solutions using a unified teacher model, Qwen3-30B-A3B, in non-thinking mode for short-CoT and thinking mode for long-CoT, ensuring consistency across methods. We evaluated all methods in both \textbf{short-CoT}, where reasoning traces are appended directly before the final answer~\cite{wei2022chain}, and \textbf{long-CoT}, where models are guided to perform detailed reasoning, including intermediate planning and possible self-reflection, before producing the final chain-of-thought. For evaluation, we use Qwen2.5-7B-Instruct and Qwen3-8B (non-thinking mode) in the short-CoT setting, and Qwen3-8B (thinking mode) and DeepSeek-R1-Distill-Qwen-7B in the long-CoT setting, representing state-of-the-art models at their scale.

\subsection{Implementation Details}
\label{imple_details}

For the MathSmith problem generation model based on Qwen3-8B, we conduct supervised fine-tuning on 8K cold-start samples generated by GPT-4o. We apply LoRA with rank 16 and train for 5 epochs on $8 \times$H100 GPUs. This stage ensures the model learns the fundamental rationale structure and format patterns of problem synthesis, providing a stable foundation for subsequent reinforcement learning.

During reinforcement learning, we adopt the verl library~\cite{sheng2024hybridflow} for policy optimization. For the structural reward defined in Eq.~\eqref{structural_reward}, we set weights $\alpha_{\text{format}} = 0.7$ and $\alpha_{\text{step}} = 0.3$ to emphasize format consistency. For the reasoning reward in Eq.~\eqref{reasoning_reward}, we set $\beta_{\text{complexity}} = 0.7$ and $\beta_{\text{consistency}} = 0.3$ to focus on generating problems that require deeper reasoning. For both complexity and consistency estimation, we set the number of teacher samples $K = 5$. We train the model using the GRPO algorithm, executing both teacher inference and MathSmith policy updates on $20 \times$H100 GPUs. The final model is selected at step 100, where performance is near convergence.

To ensure reproducibility, we adopt supervised fine-tuning for evaluation baselines using the LlamaFactory~\cite{zheng2024llamafactory} training scripts. All models are trained for 5 epochs with a learning rate of $1\mathrm{e}{-5}$, and evaluated on $8 \times$H100 GPUs under consistent settings.

\section{Results and Analysis}
\label{result_and_analysis}

\subsection{Overall Performance}
\label{exp_main}

The overall results are presented in Table~\ref{tab:main_resuilt}, revealing several key findings. Our method achieves state-of-the-art performance across multiple benchmarks, demonstrating the effectiveness of the synthesized data. We divide evaluation problems into two difficulty levels: \textit{easy \& medium} and \textit{hard}. As difficulty increases, our method yields notably stronger results, especially under the long-CoT setting, where MathSmith shows significantly larger improvements over baselines, indicating its ability to elicit more complex reasoning. Unlike prior methods that extract concepts from real math problems, MathSmith samples concept sets entirely at random, but still generalizes well to challenging real-world tasks. On certain benchmarks such as GSM8K, however, performance occasionally falls below that of the base model. This trend, also observed in other baselines, probably stems from the nature of GSM8K as a word problem data set, which differs in format and complexity from the competition-style problems emphasized during synthesis. In such cases, excessive reasoning may even hinder performance by introducing unnecessary complexity.

\subsection{Effect of Dataset Scaling}
\label{exp_data_scaling}

\begin{figure}[htbp]
\centering
\includegraphics[width=0.75\columnwidth]{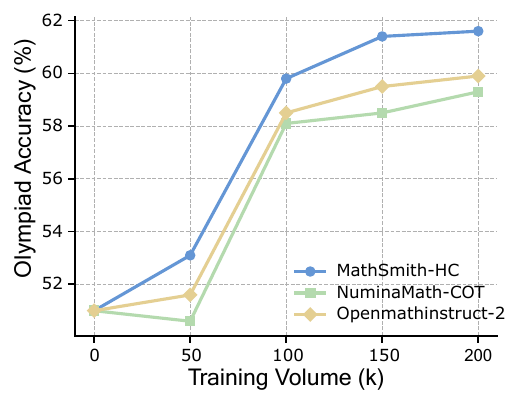}
\caption{Performance on the Olympiad benchmark under varying training data volumes (50K–200K). MathSmith-HC scales more effectively than baseline methods.}
\label{dataset_scaling_res}
\end{figure}

We evaluate the scalability of our method using the Olympiad benchmark, selected for its high difficulty, diverse problem types, and sufficient data volume, making it a more reliable indicator of performance. Qwen3-30B-A3B serves as the teacher model to synthesize all training data, while Qwen3-8B is used as the base model. As shown in Figure~\ref{dataset_scaling_res}, MathSmith-HC consistently outperforms strong baselines (NuminaMath-COT and OpenMathInstruct-2) across training sizes from 50K to 200K, with the performance gap widening as the data volume increases.

\subsection{Effect of Model Scaling}
\label{exp_model_scaling}

\begin{figure}[htbp]
\centering
\includegraphics[width=0.9\columnwidth]{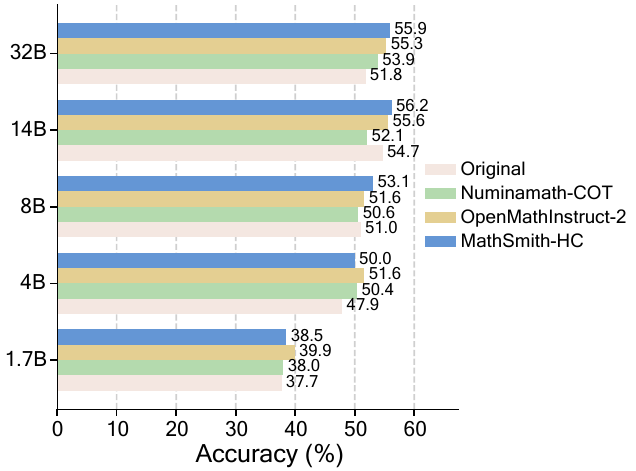}
\caption{Accuracy on the Olympiad benchmark across Qwen3 model series. MathSmith-HC yields greater gains with larger models.}
\label{model_scaling_res}
\end{figure}

We analyze the impact of model scale using the Qwen3 series, trained on a fixed 50K dataset synthesized by Qwen3-30B-A3B. As shown in Figure~\ref{model_scaling_res}, MathSmith-HC performs slightly worse on smaller models (1.7B, 4B), likely due to limited capacity to learn complex problems. As model size increases, our method consistently outperforms baselines, suggesting larger models benefit more from high-difficulty synthetic data by acquiring deeper reasoning abilities.

\subsection{Analysis of Problem Difficulty}
\label{difficult-analysis}

We assess the relative difficulty of the problems by measuring the average token length of reasoning traces generated in the thinking mode of Qwen3-30B-A3B. Following the assumption that more complex problems induce longer reasoning sequences, we collect 50 problems from each major open-source math dataset, including both training and evaluation sets, and use the model to solve them in long-CoT mode. For datasets containing fewer than 50 problems, all available instances are included. As shown in Figure~\ref{cotlength-cal}, MathSmith-SFT already produces challenging problems, indicating the effectiveness of synthesizing questions directly from difficult concepts and predefined difficulty strategies. Notably, MathSmith-HC and MathSmith-Hard result in the longest reasoning traces across all datasets, suggesting that our reinforcement learning stage further enhances problem complexity and encourages deeper reasoning behavior.

\subsection{Impact of Weakness-Focused Problem Generation}
\label{exp_self-weakness}

To evaluate the effectiveness of weakness-focused improvement, we first use MathSmith-HC to generate 10 concept-guided variants for each question in the Practice Set, targeting concepts associated with incorrect predictions. We then generate solutions to these variant problems using Qwen2.5-Math-7B, and evaluate the accuracy of a base model (Qwen2.5-Math-1.5B) on both the original and variant problems. As shown in Table~\ref{tab:self-weakness-analyse}, the weakness-focused variants yield consistent accuracy improvements over the random sampling baseline with the same number of generated problems (Epoch 1), particularly on harder problems. Furthermore, we observe that accuracy gains on the Practice Set correlate with improved generalization to other math benchmarks, suggesting that MathSmith-generated problems possess strong transferability.

\begin{table}[htbp]
\centering
\resizebox{\columnwidth}{!}{%
\begin{tabular}{cccc}
\hline
Sample   Method & Ave. (Easy \& Medium) & Ave. Hard & Acc on Practice \\ \hline
Original & 38.2 & 14.5 & 23.6 \\
WF Epoch 1 & 69.9 & 18.8 & 33.1 \\
WF Epoch 2 & 77.3 & 21.5 & 34.6 \\
WF Epoch 3 & 77.6 & 21.6 & 34.7 \\
Random & 69.4 & 15.6 & 30.0 \\ \hline
\end{tabular}%
}
\caption{Effect of weakness-focused problem generation vs. random sampling, \textbf{WF} denote Weakness-Focused pipeline.}
\label{tab:self-weakness-analyse}
\end{table}

\subsection{Ablation Analysis}
\label{ablation}

We perform ablation studies to assess the effects of different training stages in MathSmith. To quantify the usability of the problem, we define an Available Ratio, the percentage of generated problems that are correctly formatted and solvable by the teacher model with a valid answer.

After generating 50k problems for each variant, we fine-tune Qwen3-8B using the same supervised procedure. As shown in Table~\ref{tab:ablation_model_stage}, MathSmith-HC achieves the highest Available Ratio while maintaining strong performance on hard problems. Although MathSmith-Hard shows slightly better accuracy, its lower usability makes it less suitable for large-scale synthesis. We further evaluate the impact of different teacher models. Using Qwen2.5-32B to generate solutions under the Short-CoT setting, we fine-tune Qwen3-8B on 40K training samples from the questions of each method. Table~\ref{tab:ablation-exp-other-model} shows that MathSmith-HC remains consistently superior across all benchmarks.

\begin{table}[htbp]
\centering
\resizebox{\columnwidth}{!}{%
\begin{tabular}{cccc}
\hline
Training Stage & Ave. (Easy \& Medium) & Ave. Hard & Available Ratio \\ \hline
MathSmith-SFT & 87.7 & 30.3 & 71.50\% \\
MathSmith-Hard & \textbf{89.25} & \textbf{36.6} & 84.92\% \\
MathSmith-HC & 88.65 & \textbf{36.6} & \textbf{95.38\%} \\ \hline
\end{tabular}%
}
\caption{Performance of training stages. MathSmith-HC offers the best balance of accuracy and usability.}
\label{tab:ablation_model_stage}
\end{table}

\begin{table}[htbp]
\centering
\resizebox{\columnwidth}{!}{%
\begin{tabular}{ccccc}
\hline
Method & GSM8K & MATH-500 & AIME2024 & Olympiad \\ \hline
Numinamath & 92.1 & 76.0 & \textbf{23.3} & \textbf{43.8} \\
OpenMathInstruct & 91.3 & 78.2 & 13.3 & 42.3 \\
PromptCOT & 90.9 & 73.8 & 16.7 & 41.1 \\
MathSmith-HC & \textbf{93.1} & \textbf{78.8} & \textbf{23.3} & \textbf{43.8} \\ \hline
\end{tabular}%
}
\caption{Effect of different teacher models. MathSmith-HC outperforms all baselines with Qwen2.5-32B as the solver.}
\label{tab:ablation-exp-other-model} 
\end{table}

\section{Conclusion}

We introduced MathSmith, a framework for synthesizing high-difficulty mathematical problems from randomly sampled concept–explanation pairs, guided by predefined difficulty strategies and optimized via reinforcement learning for structural validity, reasoning depth, and answer consistency. By encouraging longer reasoning traces as a heuristic for problem complexity, MathSmith produces diverse and challenging problems that substantially improve LLM reasoning on Olympiad-level benchmarks. These results highlight the potential of scalable synthetic data to drive deeper reasoning capabilities in large models. Moreover, the framework demonstrates strong generalization, showing that fully synthetic problems can effectively complement limited human-authored datasets in advancing mathematical reasoning. Future work will focus on refining difficulty estimation, expanding domain coverage, and exploring adaptive generation strategies to construct richer synthetic datasets to advance high-level mathematical reasoning.

\bibliography{aaai2026}

\newpage

\appendix
\section{Appendix}
\subsection{Concept and Explanation Examples}
\label{appendix:concept_and_exp}

We collect a set of approximately 11k concept--explanation pairs from PlanetMath, following the extraction instruction described in Appendix~\ref{appendix:instruction_for_collect} and using the GPT-4o model.  
Each entry contains the mathematical concept, its concise explanation, and the associated categories.  
Table~\ref{tab:concept_examples} presents several examples.

\subsection{Training Details}

\begin{figure}[htbp]
    \centering
    \includegraphics[width=0.95\linewidth]{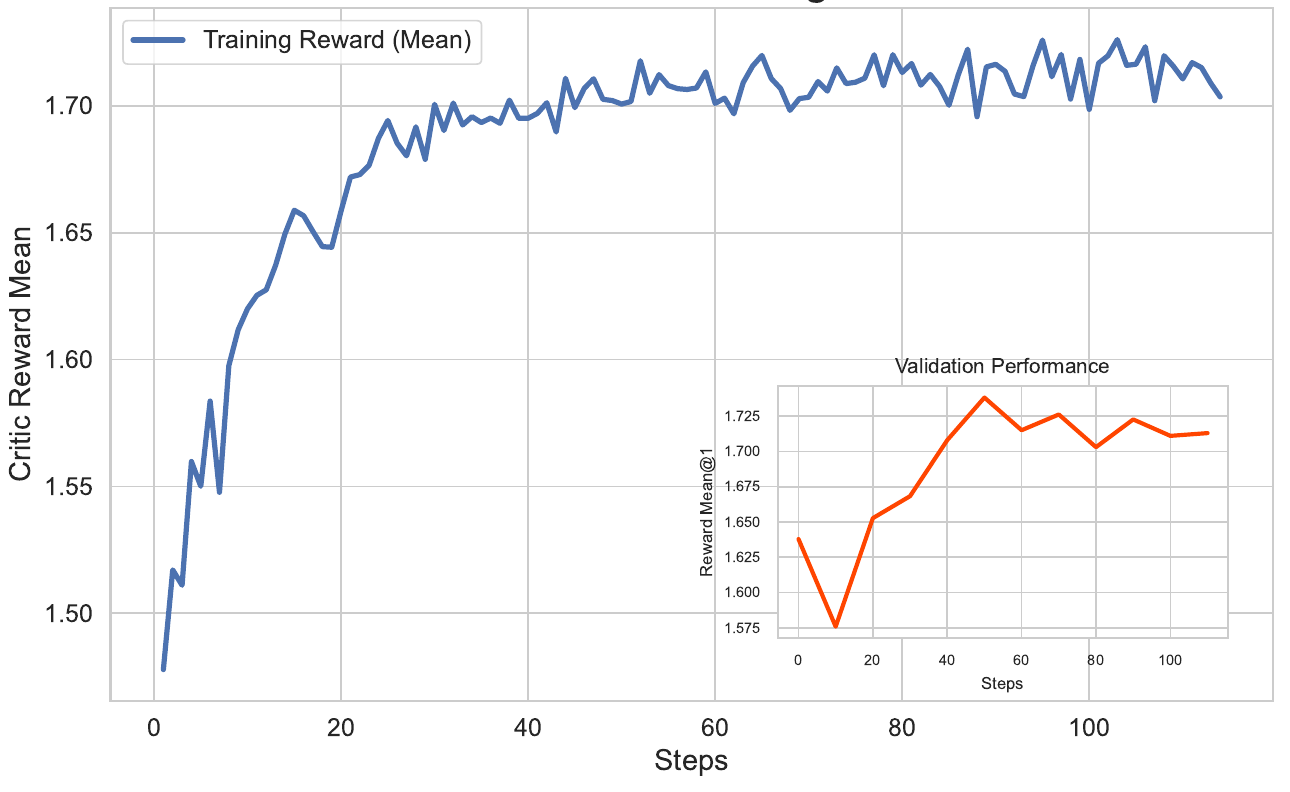}
    \caption{Reward trend during GRPO training of \textbf{MathSmith-HC}. The model steadily improves and reaches convergence after about 100 steps.}
    \label{fig:mathsmith-hc}
\end{figure}

\begin{figure}[htbp]
    \centering
    \includegraphics[width=0.95\linewidth]{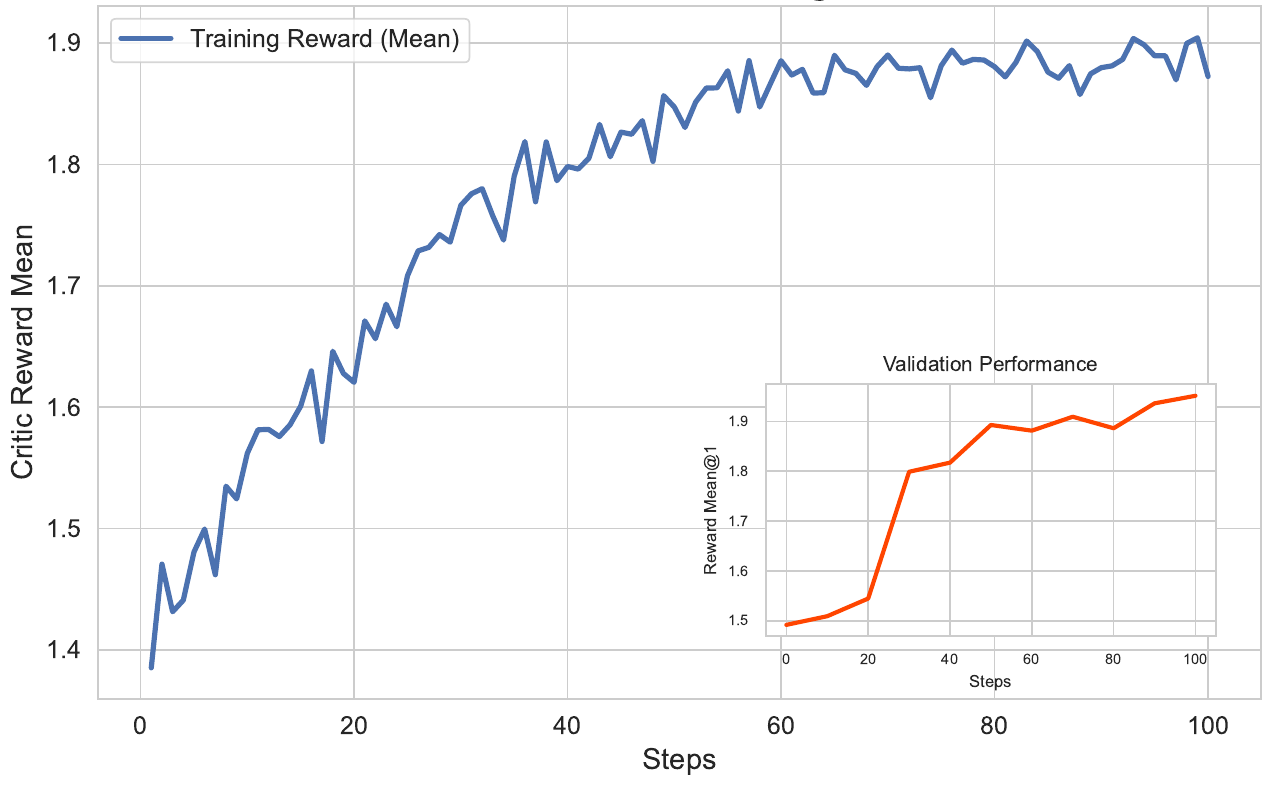}
    \caption{Reward trend during GRPO training of \textbf{MathSmith-Hard}, showing similar convergence behavior after roughly 100 steps.}
    \label{fig:mathsmith-hard}
\end{figure}

\subsection{Reinforcement Learning Training Process}
\label{appendix:rl_training}

We train \textbf{MathSmith-HC} and \textbf{MathSmith-Hard} using the GRPO algorithm within the verL framework.  
The reinforcement learning process optimizes the model through multi-objective rewards that balance structural correctness, reasoning depth, and solution consistency.  
The reward curves during training are shown in Fig.~\ref{fig:mathsmith-hc} and Fig.~\ref{fig:mathsmith-hard}.  
Both models exhibit a steady improvement in the early stages and achieve approximate convergence after around 100 steps, indicating stable reward optimization.

\subsection{Instruction Details}
\subsection{Instruction for Collect Concept and Explanation}
\label{appendix:instruction_for_collect}

\begin{tcolorbox}[title=Concept and Explanation Extraction Prompt, coltitle=white, colback=blue!3!white, colframe=blue!80!black, colbacktitle=blue!80!black, sharp corners=southwest, boxrule=0.8pt]
\small
You are a mathematical assistant.\\
Given the following markdown content from a mathematical article, extract key concepts and their explanations.

\medskip
\textbf{Instructions:}
\begin{enumerate}
  \item If the content \textbf{does NOT} define or discuss a math concept, respond with exactly: \texttt{Not a math concept}
  \item Otherwise, identify all important math concepts mentioned.
    \begin{itemize}
      \item For each concept, ensure it is explicitly mentioned or defined in the given markdown article, and provide a brief explanation based strictly on the article content.
      \item The explanation should be concise and relevant to the concept.
      \item Concept and Explanation must be written in \LaTeX-compatible format, using proper mathematical notation where appropriate.
      \item Additionally, list 1--3 broad mathematical categories related to the content.
    \end{itemize}
  \item For each concept, output a JSON format:
[
  {"Concept": "...", "Explanation": "...", "Categories": ["...", "..."]},
  ...
]
\end{enumerate}

\textbf{Input:} Markdown Math Content (from PlanetMath)
\end{tcolorbox}

\begin{table*}[htp]
\centering
\caption{Examples of Concept and Explanation pairs.}
\label{tab:concept_examples}
\renewcommand{\arraystretch}{1.0}
\begin{tabular}{p{3.5cm} p{7.5cm} p{4.5cm}}
\toprule
\textbf{Concept} & \textbf{Explanation} & \textbf{Categories} \\
\midrule

Von Neumann ordinals &
Von Neumann ordinals are defined as sets of all natural numbers less than a given number, establishing a foundation for defining natural numbers through set theory. &
Set Theory; Ordinal Numbers; Foundations of Mathematics \\

\midrule

Dynkin system &
A Dynkin system on a set $\Omega$ is a collection $\mathcal{D} \subset \mathcal{P}(\Omega)$ satisfying:
\begin{enumerate}\itemsep0pt
    \item $\Omega \in \mathcal{D}$;
    \item if $A,B \in \mathcal{D}$ and $A \subset B$, then $B \setminus A \in \mathcal{D}$;
    \item if $A_n \in \mathcal{D}$ with $A_n \subset A_{n+1}$ for $n \ge 1$, then $\bigcup_{k=1}^{\infty} A_k \in \mathcal{D}$.
\end{enumerate} &
Set Theory; Measure Theory; Probability Theory \\

\midrule

Inverse image of a function &
For a function $f$ and a subset $B \subseteq Y$, the inverse image of $B$ under the restriction ${f|}_{A}$ is  
\[
({f|}_{A})^{-1}(B) = A \cap f^{-1}(B).
\] &
Functions; Set Theory; Mathematical Analysis \\

\bottomrule
\end{tabular}
\end{table*}

\subsection{Instruction for Cold Start Data Generation}
\label{appendix:instruction_for_cold}

\begin{figure*}[htp]
\begin{tcolorbox}[title=Cold Start Data Generation Prompt with GPT-4o, coltitle=white, colback=blue!3!white, colframe=blue!80!black, colbacktitle=blue!80!black, sharp corners=southwest, boxrule=0.8pt]
\small
Given the Concepts and Explanations along with the instructions below, develop a \textbf{single challenging mathematics problem} suitable for advanced Olympiads.

\medskip
\textbf{Instructions:}

\textbf{A. Select Relevant Concepts}
\begin{enumerate}
  \item Carefully analyze the provided concepts and their explanations.
  \item Based on this analysis, choose a suitable subset of concepts (at least two; more are welcome if appropriate) that can be naturally and meaningfully integrated into a cohesive, well-focused mathematics problem.
\end{enumerate}

\textbf{B. Problem Difficulty Rules}
Include \textbf{at least two} of the following features:
\begin{itemize}
  \item \textbf{Multi-step Reasoning:} Requires multiple sequential logical steps.
  \item \textbf{Cross-topic Integration:} Combines distinct mathematical topics.
  \item \textbf{Implicit or Reverse Logic:} Includes hidden conditions or reverse deduction.
  \item \textbf{Distractors:} Contains misleading or extraneous conditions.
  \item \textbf{Abstract Modeling:} Translates complex scenarios into mathematical form.
  \item \textbf{Multiple Solution Paths:} Allows various non-trivial solving methods.
  \item \textbf{Advanced Manipulation:} Necessitates sophisticated algebraic or geometric transformations.
  \item \textbf{Extreme Conditions:} Focuses on limits or boundary values.
  \item \textbf{Non-standard Representation:} Uses unconventional presentation of familiar concepts.
\end{itemize}

\textbf{C. Problem Format Constraints}
\begin{enumerate}
  \item Generate \textbf{only one single problem}; no sub-questions or multiple parts.
  \item The problem must have a \textbf{single, unified solving goal}, clearly answerable by a focused line of reasoning. Avoid phrases like ``first..., then...'' or numbering subtasks.
  \item The problem must admit a \textbf{well-defined, verifiable solution}, such as a number, expression, or equation.
  \item \textbf{Do not include proof-based phrasing} (e.g., ``prove that,'' ``show that''). Ask only for specific, verifiable results.
  \item Avoid secondary tasks (e.g., ``construct an example,'' ``justify your answer'') unless construction is the core objective. \textbf{Keep the question focused}.
\end{enumerate}

\medskip
\textbf{Output Format:}

\textbf{A. Rationale}  
Explain your thought process step-by-step:
\begin{enumerate}
  \item Analyze and understand the given concepts in depth.
  \item Select a suitable combination of concepts that can be meaningfully integrated.
  \item Explain how these concepts are woven into a unified mathematical scenario.
  \item Identify and describe the difficulty features used.
  \item Formulate the final problem statement clearly and concisely.
\end{enumerate}
Wrap your rationale in:
\begin{verbatim}
<rationale>
[Your construction thought process goes here]
</rationale>
\end{verbatim}

\textbf{B. Problem}  
Present the final problem inside:
\begin{verbatim}
<problem>
[Your final math problem goes here]
</problem>
\end{verbatim}

\textbf{Given Concept and Explanation:} \verb|{CONCEPT_AND_EXPLANATION}|
\end{tcolorbox}
\end{figure*}

\subsection{Instruction for Quetion and Solution Generation}
\label{appendix:instruction_for_qag}

\begin{figure*}[htp]
\begin{tcolorbox}[title=Concept-Based Question and Solution Generation Prompt, coltitle=white, colback=blue!3!white, colframe=blue!80!black, colbacktitle=blue!80!black, sharp corners=southwest, boxrule=0.8pt]
\small
Given the Concepts and Explanations along with the instructions below, develop a \textbf{single challenging mathematics problem} suitable for advanced Olympiads.

\medskip
\textbf{Instructions:}

\textbf{A. Select Relevant Concepts}
\begin{enumerate}
  \item Carefully analyze the provided concepts and their explanations.
  \item Choose a subset of concepts that can be naturally and meaningfully integrated into a single, well-focused problem.
\end{enumerate}

\textbf{B. Problem Difficulty Rules}
\begin{itemize}
  \item The problem should require deep insight and non-obvious reasoning.
  \item Include \textbf{at least two} of the following difficulty features:
    \begin{itemize}
      \item \textbf{Multi-step Reasoning}: Requires multiple sequential logical steps.
      \item \textbf{Cross-topic Integration}: Combines distinct mathematical topics.
      \item \textbf{Implicit or Reverse Logic}: Involves hidden constraints or backward deduction.
      \item \textbf{Distractors}: Includes misleading or irrelevant elements.
      \item \textbf{Abstract Modeling}: Converts complex real-world scenarios into mathematical expressions.
      \item \textbf{Multiple Solution Paths}: Allows for multiple viable solving strategies.
      \item \textbf{Advanced Manipulation}: Requires sophisticated algebraic or geometric techniques.
      \item \textbf{Extreme Conditions}: Focuses on limits, edge cases, or extremal values.
      \item \textbf{Non-standard Representation}: Presents concepts in unconventional forms.
    \end{itemize}
\end{itemize}

\textbf{C. Output Format}
Strictly format your response as:

\begin{verbatim}
<rationale>
Step-by-step reasoning describing:
  - How the selected concepts are connected;
  - How they contribute to the structure and difficulty of the problem;
  - Which difficulty features (from B.2) are used, and how they are embedded.
</rationale>

<problem>
[Write a single self-contained, high-level Olympiad mathematics problem here.]
</problem>
\end{verbatim}

\textbf{Given Concept and Explanation:} \verb|{Sampled_Concept_and_Explanation_Set}|

\medskip
\textbf{Solution Prompt Instruction:}
\begin{verbatim}
Please reason step by step, and put your final answer within \boxed{}.
\end{verbatim}
\end{tcolorbox}
\end{figure*}

\subsection{Case Study}

We present illustrative examples of the problem generation processes for MathSmith-HC and MathSmith-Hard. 
MathSmith-HC offers more detailed rationale steps, enhancing interpretability, whereas MathSmith-Hard generates concise rationales that still yield readable problems.

\begin{figure*}[htbp]
\begin{tcolorbox}[colback=white, colframe=black, sharp corners, boxrule=0.8pt, enhanced]
\textbf{Sampled Concept and Explanation:}\\
1.\textbf{Quadratic Resolvent}: The quadratic resolvent of the cubic equation $y^{3} + py + q = 0$ is the auxiliary equation $z^{2} + qz - \left( \frac{p}{3} \right)^{3} = 0$, which helps to find roots $z_{1} = u^{3}$ and $z_{2} = v^{3}$, where $y = u + v$ satisfies the original cubic equation.\\
2.\textbf{GCD condition}: The condition $\gcd(a,b)=1$ can be added to the conclusion of Dirichlet's approximation theorem.\\
3.\textbf{Partial Map}: A partial map on a set $X$ is an application defined from a subset of $X$ into $X$, denoted as ${\mathfrak{F}}(X)$.\\
4.\textbf{Minimum Element}: A minimum element of a set $S$ is the smallest element in $S$. If $S$ is a finite subset of $\mathbb{R}$, then it contains a minimum element $s \in S$.\\
5.\textbf{Lattices with Operators}: Lattices that include additional operations beyond the standard meet and join.

\textbf{Rationale:}

\textbf{Step 1: Analyze and understand the given concepts in depth.}

The concepts provided span algebra (quadratic resolvent of cubic equations), number theory (GCD conditions), set theory (partial maps and minimum elements), and abstract algebra (lattices with operators). Among these, quadratic resolvent and minimum element naturally lend themselves to integration with algebraic structures, while GCD conditions and partial maps introduce constraints or mappings that can add complexity.


\textbf{Step 2: Select a suitable combination of concepts.}

I chose quadratic resolvent and minimum element as the primary concepts, integrating them with constraints inspired by GCD conditions. This combination allows for a rich interplay between algebraic manipulation and optimization under specific conditions.


\textbf{Step 3: Weave concepts into a unified scenario.}

The problem is constructed around finding roots of a cubic equation under constraints tied to GCD conditions and minimum elements. By embedding implicit conditions and requiring advanced algebraic manipulation, the problem challenges competitors to navigate multiple layers of reasoning.


\textbf{Step 4: Incorporate difficulty features.}\\
--\textbf{Multi-step reasoning}: Solving the cubic equation requires sequential steps involving resolvent construction, GCD constraints, and optimization.\\
--\textbf{Implicit logic}: Hidden conditions tied to GCD and minimum elements force competitors to deduce constraints indirectly.\\
--\textbf{Advanced manipulation}: Algebraic transformations of the cubic resolvent and interaction with GCD constraints demand sophisticated handling.\\
--\textbf{Extreme conditions}: The problem focuses on minimizing a specific expression under strict constraints, pushing competitors to consider boundary cases.


\textbf{Step 5: Formulate the final problem statement.}

The problem is presented concisely, ensuring competitors focus on a single, unified goal without auxiliary subtasks.

\textbf{Final Problem:}

Let \( y^3 - 3py + q = 0 \) be a cubic equation with integer coefficients \( p \) and \( q \), 
where \( p > 0 \) and \( q < 0 \). 
The quadratic resolvent of this equation is given by 
\( z^2 + qz - \left( \frac{p}{3} \right)^3 = 0 \), 
with roots \( z_1 \) and \( z_2 \). 
Suppose \( y = u + v \) satisfies the original cubic equation, 
where \( u^3 = z_1 \) and \( v^3 = z_2 \).
Additionally, let \( u \) and \( v \) satisfy the following conditions:
1. \( \gcd(u, v) = 1 \),
2. \( u + v \) is the minimum positive real root of \( y^3 - 3py + q = 0 \).
If \( p = 9 \) and \( q = -27 \), determine the exact value of \( u^2 + v^2 \).

\end{tcolorbox}
\caption{Case study of MathSmith-HC problem generation.}
\end{figure*}


\begin{figure*}[htbp]
\begin{tcolorbox}[colback=white, colframe=black, sharp corners, boxrule=0.8pt, enhanced]
\textbf{Sampled Concept and Explanation:}
\begin{enumerate}
    \item \textbf{Analytic Polyhedron:} 
    An analytic polyhedron is defined as the set 
    $\Omega := \{ z \in W \mid |f_j(z)| < 1, \, j = 1,\ldots,k \}$ 
    which is relatively compact in an open set $W$ of a domain 
    $G \subset \mathbb{C}^{n}$, where $f_1,\ldots,f_k$ are holomorphic functions.
    \item \textbf{Modulus of a Complex Number:} 
    The modulus of a complex number is the distance from the origin in the complex plane, denoted as $r$.
    \item \textbf{Infinitesimal:} 
    An infinitesimal is a number that is greater than zero but smaller than any positive real number, 
    defined such that for all positive integers $n$, $|\varepsilon| < \frac{1}{n}$.
    \item \textbf{Graph Isomorphism:} 
    Graph isomorphism refers to a one-to-one correspondence between the vertex sets of two graphs 
    that preserves the edge connectivity.
    \item \textbf{n-th Derivative of a Determinant:} 
    The n-th derivative of the determinant of a matrix $A$, 
    where $A$ is a $d \times d$ matrix with entries as functions of $t$, 
    is expressed using a sum over multinomial coefficients and permutations, 
    involving the derivatives of the matrix entries.
\end{enumerate}

\vspace{0.5em}
\textbf{Rationale:}

\textbf{Step 1. Analyze the given concepts:}

The concepts involve analytic geometry, complex numbers, infinitesimals, graph theory, and matrix calculus. These are diverse, allowing for cross-topic integration.

\vspace{0.5em}

\textbf{Step 2. Select suitable concepts:}

Combine Modulus of a Complex Number and n-th derivative of a determinant, as they can naturally interact in a problem involving complex-valued matrices and constraints.

\vspace{0.5em}

\textbf{Step 3. Integration:}

Construct a scenario involving a complex matrix whose entries are constrained by modulus conditions, requiring advanced manipulation of determinants and derivatives.

\vspace{0.5em}

\textbf{Step 4. Difficulty features:}

Incorporate multi-step reasoning, cross-topic integration, and implicit logic by embedding hidden modulus constraints and derivative conditions.

\vspace{0.5em}

\textbf{Step 5. Formulate problem:}

Design a focused problem requiring determinant analysis under complex modulus restrictions and derivative evaluation.

\vspace{0.5em}
\textbf{Final Problem:}

Let $A(t)$ be a $3 \times 3$ complex matrix whose entries are given by
\[
    a_{ij}(t) = t^{i+j-2} + \frac{\omega_{ij}}{1 + |t|^2},
\]
where $\omega_{ij}$ are distinct complex numbers with $|\omega_{ij}| = 1$ for all 
$i, j \in \{1, 2, 3\}$, and $t \in \mathbb{C}$ satisfies $|t| < 1$.

Define $D(t) = \det(A(t))$, the determinant of $A(t)$.  
If the second derivative $D''(t)$ vanishes at exactly two distinct values 
$t_1, t_2 \in \mathbb{C}$ with $|t_1| < |t_2| < 1$, determine the sum
\[
    S = |D(t_1)| + |D(t_2)| + \sum_{i,j=1}^3 |\omega_{ij}|^3.
\]
\end{tcolorbox}
\caption{Case study of MathSmith-Hard problem generation.}
\end{figure*}

\end{document}